\documentclass[10pt,twocolumn,letterpaper]{article}

\usepackage{cvpr}              
\usepackage{multirow}
\usepackage[accsupp]{axessibility} 
\usepackage[dvipsnames]{xcolor}

\definecolor{cvprblue}{rgb}{0.21,0.49,0.74}
\usepackage[pagebackref,breaklinks,colorlinks,citecolor=cvprblue]{hyperref}
\def\paperID{8} 
\def\confName{CVPR}
\def\confYear{2024}

\title{Efflex: Efficient and Flexible Pipeline for Spatio-Temporal Trajectory Graph Modeling and Representation Learning}

\author{
    Ming Cheng$^{1*}$, 
    Ziyi Zhou$^{1*}$, 
    Bowen Zhang$^{2}$\thanks{Equal contribution.} , 
    Ziyu Wang$^3$,
    Jiaqi Gan$^1$,
    Ziang Ren$^1$,\\
    Weiqi Feng$^4$,
    Yi Lyu$^5$,
    Hefan Zhang$^1$,
    Xingjian Diao$^1$\thanks{Corresponding author.}
    \\
    $^1$Dartmouth College
    $^2$Shanghai Jiao Tong University
    $^3$University of California, Irvine \\
    $^4$Harvard University
    $^5$Independent Researcher
    \\
    \tt \small \{ming.cheng.gr, ziyi.zhou.gr, xingjian.diao.gr\}@dartmouth.edu
}
\begin{document}
\maketitle
\begin{abstract}
In the landscape of spatio-temporal data analytics, effective trajectory representation learning is paramount. 
To bridge the gap of learning accurate representations with efficient and flexible mechanisms, 
we introduce Efflex, a comprehensive pipeline for transformative graph modeling and representation learning of the large-volume spatio-temporal trajectories. Efflex pioneers the incorporation of a multi-scale k-nearest neighbors (KNN) algorithm with feature fusion for graph construction, marking a leap in dimensionality reduction techniques by preserving essential data features. 
Moreover, the groundbreaking graph construction mechanism and the high-performance lightweight GCN increase embedding extraction speed by up to 36 times faster. We further offer Efflex in two versions, Efflex-L for scenarios demanding high accuracy, and Efflex-B for environments requiring swift data processing. Comprehensive experimentation with the Porto and Geolife datasets validates our approach, positioning Efflex as the state-of-the-art in the domain. Such enhancements in speed and accuracy highlight the versatility of Efflex, underscoring its wide-ranging potential for deployment in time-sensitive and computationally constrained applications.
\end{abstract}    
\section{Introduction}
\label{sec:intro}

\begin{figure}[htbp]
  \centering
    \includegraphics[width=\linewidth]{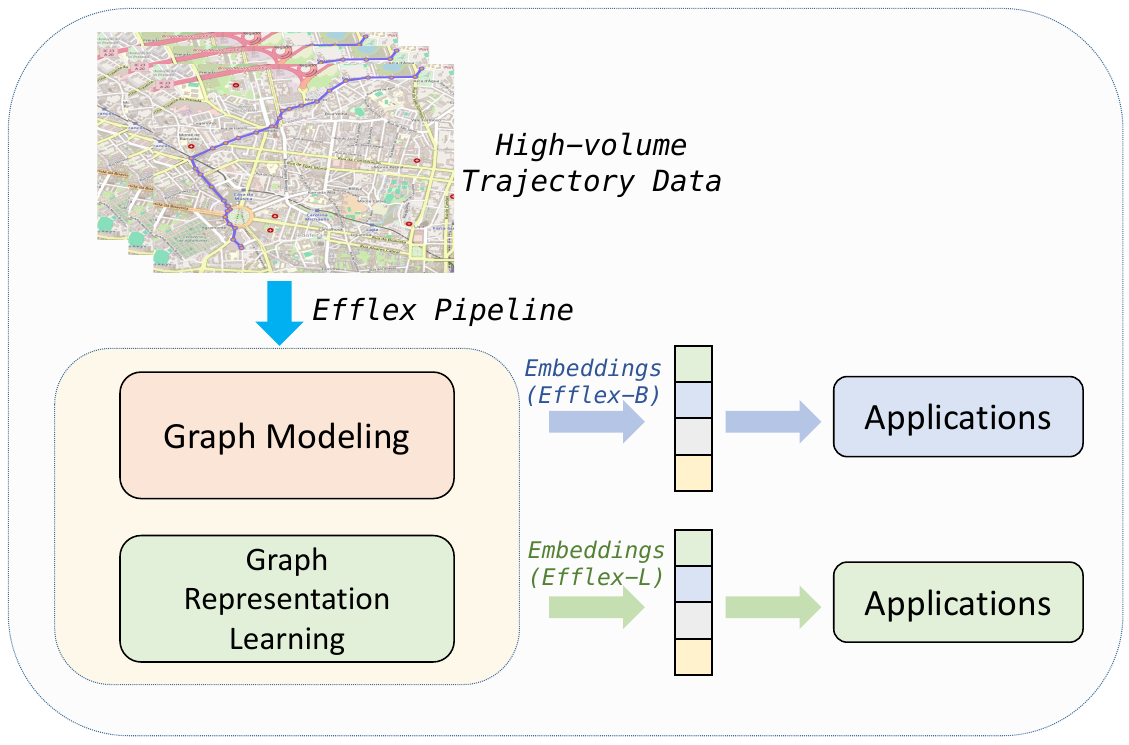}
    \caption{
    \textbf{The proposed Efflex pipeline.}
    We offer two models, Efflex-B and Efflex-L, to learn accurate embeddings from the original high-volume trajectory data.
    Efflex-B specializes in improving the speed while Efflex-L focuses on obtaining state-of-the-art performance, indicating different applications for each model. 
    }
  \label{fig:teaser}
\end{figure}

The analysis of high-volume spatio-temporal data, which captures both the location of human activities and their movement over time \cite{wang2020deep}, is becoming increasingly essential across various fields including cloud computing \cite{zhang2023first,zhang2021tapping, yao2020privacy}, recommender systems \cite{yin2016spatio, huang2022empowering}, network management \cite{yang2022zebra}, smart healthcare \cite{wang2020guardhealth, cheng2023saic} and monitoring \cite{wang2024differential, zhou2021doseguide, zhang2023doseformer}, social policy analysis \cite{ma2022assessing}, and localization-based services \cite{hu2020dasgil, zhu2020sparking}. However, spatio-temporal data generated from real-world human activities often come in large volumes, diverse formats, and with a lot of non-useful information, leading to significant storage and analysis expenses \cite{cheng2024vetrass}. These issues limit the data's practical use. Consequently, deriving meaningful insights from raw spatio-temporal data to create dense and informative representations has become a critical focus in the field of computer science. The essence of utilizing this data effectively lies in transforming it into embeddings -- simplified representations that highlight underlying patterns, facilitating easier analysis and prediction by computational models.

While traditional approaches to spatio-temporal data analysis have largely relied on dimensionality reduction techniques \cite{wold1987principal, wall2003singular, kruskal1978multidimensional} and neural network-based methodologies \cite{meiler2001generation, dziugaite2015neural, bengio2013representation}, these methods often grapple with limitation in processing speed, generalization to unseen data, and complex pre-processing overhead \cite{avalos2018representation, bengio2013representation}. To be specific, owing to the time-ordered characteristics of spatio-temporal data, many current techniques utilize neural architectures capable of sequence processing (including LSTM and RNN) for feature extraction and data representation learning \cite{li2018deep, yang2021towards, ding2022analyzing}. Yet, these approaches often necessitate extensive preliminary data processing, which involves organizing spatial data into grids and using pre-arranged spatio-temporal datasets. Furthermore, the intricate design of these sequential neural networks contributes to significant training demands. Graph neural networks (GNNs) have emerged as a promising alternative, offering a powerful means of spatio-temporal representation learning \cite{cheng2024vetrass, han2021graph, yao2022trajgat}. \textit{\textbf{However, there remains a significant gap: a specialized end-to-end pipeline that addresses the unique characteristics of spatio-temporal graph modeling with a focus on flexible and efficient learning mechanisms. }}

In response to this gap, we introduce Efflex, an efficient and flexible pipeline designed specifically for spatio-temporal trajectory graph modeling and representation learning, as shown in Figure \ref{fig:teaser}. Efflex leverages innovative techniques to construct graphs directly from raw trajectory data and learn from them efficiently, marking a significant departure from conventional methods. Our contributions through the development of Efflex are threefold:

\begin{itemize}
    \item \textbf{Multi-scale graph construction.} To the best of our knowledge, we are the \textbf{\textit{first}} to apply a multi-scale k-nearest neighbors (KNN) algorithm with feature fusion for graph construction, achieving nuanced dimensionality reduction while retaining essential trajectory data. Our innovation sets a new standard for capturing the complexity of spatio-temporal information.
    \item \textbf{State-of-the-art performance.} We develop a custom-built lightweight Graph Convolutional Network (GCN) that significantly enhances the model's efficiency. Compared to existing methodologies, our lightweight GCN improves embedding extraction speed up to \textbf{36 times faster} while maintaining competitive accuracy.
    \item \textbf{Generalized and flexible framework.} Efflex offers two versions tailored to diverse application needs. Efflex-L prioritizes precision with node2vec \cite{grover2016node2vec}, while Efflex-B focuses on speed with our GCN, proving the framework's adaptability and broad real-world applicability.
\end{itemize}

\section{Related Work}
\label{sec:related}

\subsection{Matrix Factorization-Based Methods}

Matrix factorization approaches are key in representation learning, typically reducing high-dimensional data into a more practical form while striving to maintain the integrity of the original data through matrix transformation \cite{mnih2007probabilistic, dziugaite2015neural, mackiewicz1993principal}.

As a prior work, PCA reduces dimension by projecting data onto a hyperplane structured to capture maximum variance, thus ensuring a robust representation of the data's original structure \cite{bengio2013representation, wold1987principal}. SVD follows a similar reduction principle but factorizes the data matrix into orthogonal components, which help isolate independent information sources within the data \cite{abu2021implicit, wall2003singular}. Meanwhile, MDS focuses on dimensional reduction by striving to conserve the pairwise distances between data points, aiming to uphold the spatial relationships post-reduction \cite{tenenbaum2000global, kruskal1978multidimensional}.

Although these methods are cornerstones of data analysis, their rigid mathematical underpinnings can lead to sub-optimal performance on sizable or intricate datasets\cite{AlikhaniKoshkak2024SEAL}. They often struggle to adapt to novel, unseen data, especially within the ever-changing contexts of real-world applications \cite{avalos2018representation, bengio2013representation}.

\subsection{Learning-Based Methods}

In recent years, learning-based methods using neural networks such as LSTM and RNN have been pivotal for efficiently learning representations from spatio-temporal data, capitalizing on their sequential dynamics \cite{li2018deep, yao2019computing, huang2023lstm}. Pei \textit{et al.} \cite{pei2016modeling} proposed Siamese Recurrent Networks (SRNs) to model time series similarities through recurrent neural networks, offering a fresh perspective on embedding learning. Similarly, NEUTRAJ \cite{yao2019computing} introduces a seed-guided neural metric learning method to efficiently compute trajectory similarities, leveraging RNNs for scalable and effective analysis. T3S \cite{yang2021t3s} combines RNNs with attention mechanisms for nuanced representation learning of trajectory data, enhancing the accuracy of similarity computations. These approaches highlight the adaptability and efficiency of learning-based models in capturing the complexities of data through advanced neural network techniques. However, while these approaches are effective in identifying temporal characteristics, their extensive resource requirements for training pose challenges for widespread application and generalization in real-world settings.

Meanwhile, the field of graph representation learning has also seen significant innovations \cite{li2015gated, grover2016node2vec, hamilton2017inductive}. GGSNN merges gated recurrent units and graph neural networks to dynamically refine node representations, thus improving the detection of complex relationships \cite{li2015gated}. Concurrently, node2vec \cite{grover2016node2vec} utilizes sophisticated random walk strategies to define and explore node neighborhoods, thereby enhancing feature learning. Further, Hamilton \textit{et al.} \cite{hamilton2017inductive} proposed GraphSAGE, which creates node embeddings by aggregating features from local neighborhoods, facilitating learning from large-scale graphs. These methods represent crucial advancements in graph analysis, however, they mainly focus on static structures, leaving a gap in capturing the spatio-temporal dynamics inherent to many real-world scenarios, highlighting the ongoing need for models that effectively integrate the spatio-temporal aspects of data.

\section{Method}
\label{sec:method}
\begin{figure*}[htbp]
  \centering
    \includegraphics[width=\textwidth]{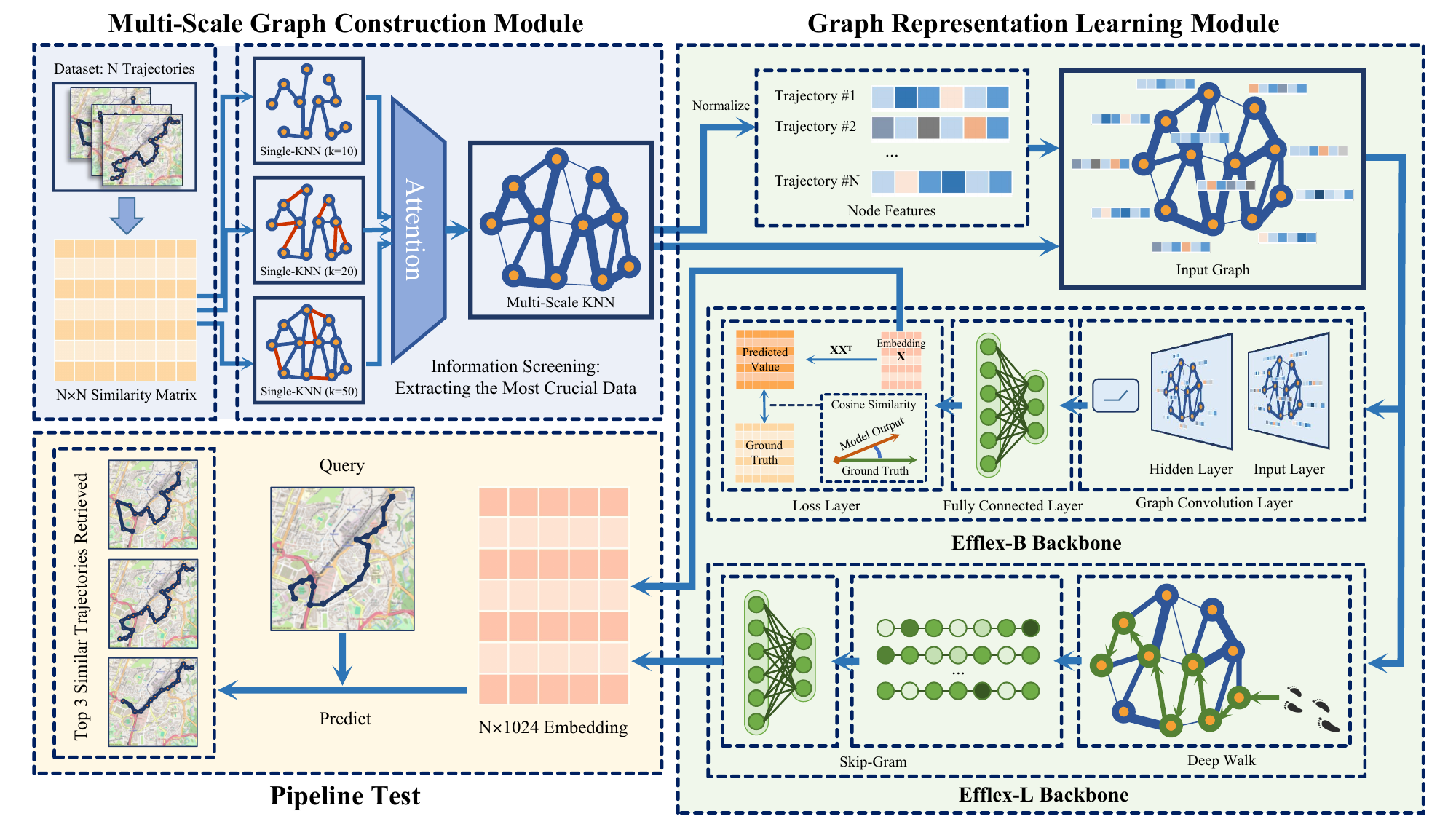}
    \caption{
    \textbf{Overview of the Efflex pipeline.} 
    \textbf{Pipeline Train:} 
    We build the graph from original trajectory data using multi-scale KNN algorithms with feature fusion by an attention module. The adjacent matrix and node features are then input into a lightweight GCN (Efflex-B) / node2vec \cite{grover2016node2vec} (Efflex-L) for accurate embedding learning. 
    Efflex-B specializes in improving the speed significantly while Efflex-L embraces state-of-the-art performance. 
    \textbf{Pipeline Test:} 
   We conduct the \textit{top-k} trajectory search experiment where given a query trajectory, the model outputs the \textit{top-k} similar ones. The precise search results indicate Efflex's ability to learn high-quality representations of the original data. 
    }
  \label{fig:pipeline}
\end{figure*}

The overview of the Efflex pipeline is shown in Figure \ref{fig:pipeline}, 
which involves two parts. The Multi-Scale Graph Construction Module specializes in constructing adjacent matrices representing edge connections within the graph, and the Graph Representation Learning Module learns accurate graph embeddings.

\subsection{Multi-Scale Graph Construction}
\label{sec:adjmatrix}
\subsubsection{Graph Construction From Trajectories}
To bridge the gap between trajectory similarity and graph topology, and convert the trajectory representation learning problem into the task of graph embedding learning, each trajectory is represented as a vertex in the graph $G(V, E, S)$, where $V$, $E$, and $S$ represent the vertex set, edge set, and weighted adjacent matrix, respectively. 

Formally, assume $\mathcal{T} = \{T_1, T_2, ..., T_n\}$ as the set of $n$ trajectories, each vertex $v_i \in V$ represents each trajectory $T_i \in \mathcal{T}$. 
Inspired by \cite{yao2019computing}, 
the connection between vertex is determined through the $k$-nearest neighbors (KNN) algorithm: If $T_i$ and $T_j$ are $k$-nearest neighbors, an edge $e_{ij} \in E$ exists.
$S = (s_{ij})_{|V|\times|V|}$, as the weighted adjacent matrix, quantitatively reflexes the edge connection between vertex, which is computed through the equation below:
\begin{equation}
\label{eq:adj}
    s_{ij} = \frac{e^{dist(T_i, T_j)}}{\sum_{T_k \in \mathcal{K}}e^{dist(T_i, T_k)}}
\end{equation}
where $\mathcal{K}$ indicates the set of $k$-nearest neighbors of trajectory $T_i$, and $dist(\cdot,\cdot)$ represents the distance function (Fréchet \cite{frechet1906quelques}, Hausdorff \cite{belogay1997calculating}, and DTW \cite{gold2018dynamic}). 
In Equation \ref{eq:adj}, $s_{ij}$ measures the weight of connections between vertex $i$ and $i$ in the graph, and the adjacent matrix $S$ of graph $G$ is computed for the certain $k$.

\begin{figure}[htbp]
  \centering
    \includegraphics[width=\linewidth]{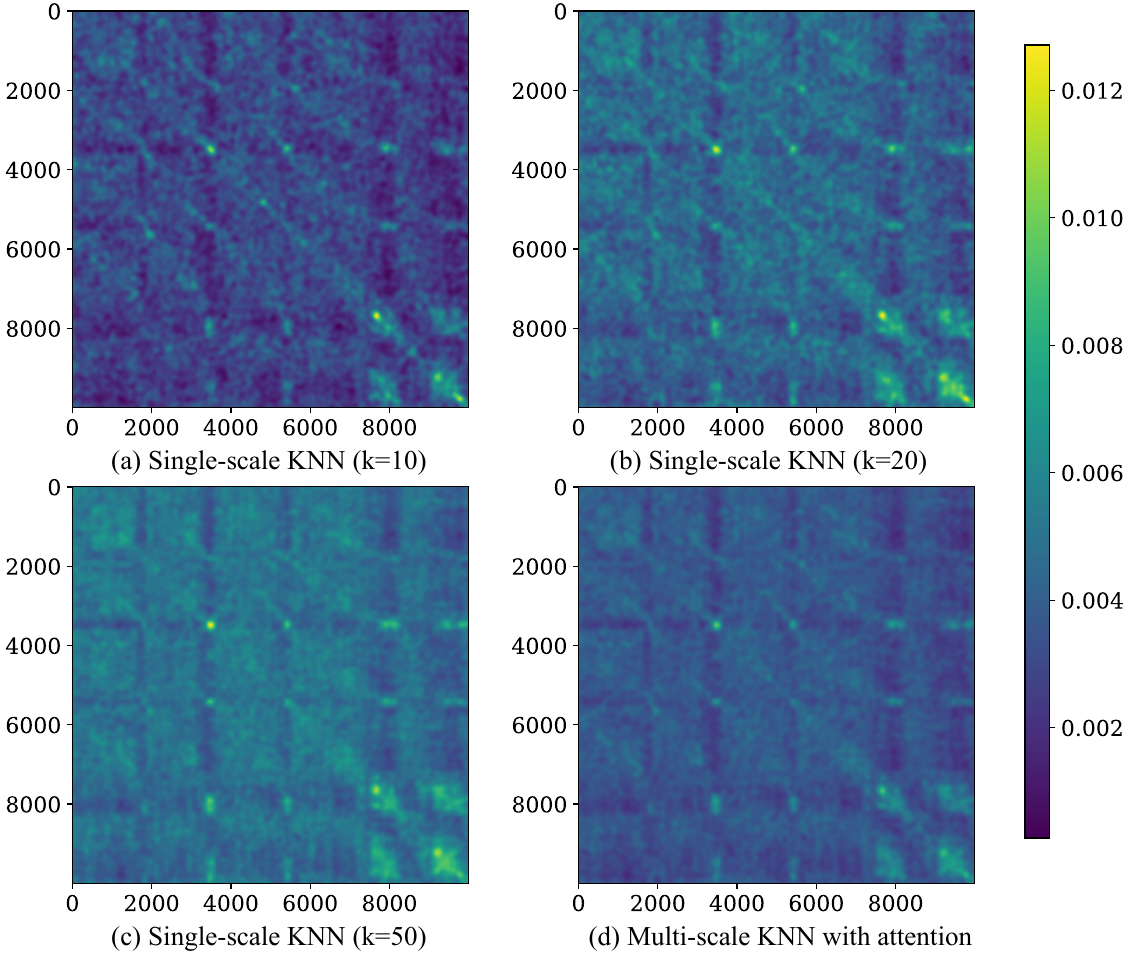}
    \caption{
    \textbf{Visualization of edge connection weights.}
    \textbf{Figure (a) - (c):} Edge connection weights obtained by single-scale KNN ($k$ = 10, 20, 50).
    \textbf{Figure (d):} Connection weights fused by multi-scale KNN with the attention mechanism.
    }
  \label{fig:attention}
\end{figure}

\subsubsection{Multi-Scale Graph Construction and Fusion}
Considering that different values of $k$ significantly affect the measure of trajectory similarity, we compute multiple adjacent matrices based on different $k$ values. 
Specifically, larger $k$ values capture more global information and lead to a comprehensive understanding of the overall graph, while smaller $k$ provide a detailed view of local connections, potentially revealing finer and localized patterns within the graph. 
Formally, for specific $k_m$, the corresponding adjacent matrix $S_m$ is obtained following the process above. Therefore, for $K = \{k_1, k_2, ..., k_m\}$, we can get:
\begin{equation}
\label{eq_s}
    \mathcal{S} = \{S_1, S_2, ..., S_m\}
\end{equation}
which represents a set of weighted adjacent matrices in multi-scale.
$\mathcal{S}$ will then be fused together for the Graph Representation Learning Module to extract graph embeddings. 

We employ a lightweight linear transformation-based attention mechanism, inspired by \cite{diao2023ft2tf, diao2023av}, to facilitate the extraction of intricate patterns and dependencies among the adjacent matrix set $\mathcal{S}$. 
Formally, assume $\mathcal{S} = \{S_1, S_2, ..., S_m\}$ in Equation \ref{eq_s} as the set of adjacency matrix, the stacked matrix is applied to a sequence of learnable linear transformations followed by non-linear activation functions to compute the attention weights $W$, as expressed below:
\begin{equation}
    W = Attn(Stack(\mathcal{S})), 
    Attn(\cdot) = Seq[LT(\cdot)f(\cdot)]
\end{equation}
where $Stack(\cdot)$ indicates the stack operation among all matrices, $Seq[\cdot]$ represents the sequential blocks, $LT(\cdot)$ and $f(\cdot)$ 
refer to the linear transformation and non-linear activation function (LeakyReLU), respectively. 
Afterward, the fused adjacent matrix $S'$ can be expressed by:
\begin{equation}
    S' = Norm(MatMul(W, Stack(\mathcal{S})))
\end{equation}
where $MatMul$ is the matrix multiplication, and $Norm(\cdot)$ indicates the normalization operation to map the connection weights in the adjacent matrix within $[0, 1]$.

Since the initial adjacent matrices are constructed with different $k$ values, where larger values aim to capture broad and global relationships of the graph while smaller values focus on extracting localized patterns among nearby nodes, this fusion procedure instructs the model to selectively leverage features in multi-scale by dynamically assigning weights to each adjacency matrix.
The qualitative visualization of edge connection weights is shown in Figure \ref{fig:attention}.
Section \ref{subsection:ablation} further proves the effectiveness of this design.

\subsection{Graph Representation Learning Module}
The Graph Representation Learning Module is mainly designed as a sequential lightweight Graph Convolutional Network (GCN) \cite{kipf2016semi, wu2019simplifying}, aiming to generate accurate graph embeddings based on input adjacent matrix and node features. 
Formally, given $F_V$ and $S'$ as the node features and adjacent matrix, the sequential GCN model $M(\theta, S', F_V)$ with trainable parameters $\theta$ can be represented as:
\begin{equation}
\begin{split}
    M(\theta, S', F_V) &= Seq[MatMul(S', \\ &MatMul(F_V, W(\theta))) 
    + \delta(\theta)]
\end{split}
\end{equation}
where $Seq[\cdot]$ represents the sequential blocks, $MatMul$ indicates the matrix multiplication, $W(\theta)$ and $\delta(\theta)$ refer to the learnable weights and bias parameters within the GCN. Node features $F_V$ for pipeline training are obtained through the adjacent matrix $S'$ with normalization operation and self-loops ($1$ on the diagonal).
The output of the model ($Em(\theta) \in \mathcal{R}^{N\times d}$) with parameters $\theta$ is the learned embedding of $N$ trajectories, each as a  $1\times d$ embedding
 vector ($d$ is the preset embedding dimension).

To instruct the model to generate accurate graph embeddings, we employ cosine similarity distance \cite{kryszkiewicz2014cosine, yao2020cosine} as the loss function and AdamW \cite{loshchilov2017decoupled} as the optimizer:
\begin{equation}
    \theta = argmin_\theta(Cosine(GT, Em(\theta)Em(\theta)^T))
\end{equation}
where $Em(\theta) = M(\theta, S', F_V)$ is the embedding generated by the model, and $GT$ indicates the ground truth by computing Euclidean distance (actual distance) between every two trajectories in the dataset. 
The learnable parameters $\theta$ will be optimized for each epoch. Ablation studies on other loss functions are shown in Section \ref{subsection:ablation}.

The designed lightweight structure allows our model to converge remarkably fast, which significantly reduces the training time compared to other existing graph representation learning models \cite{grover2016node2vec} while guaranteeing the generation of accurate embedding, as shown in Section \ref{subsec:effana}.

\begin{table*}[htb]
\centering
\resizebox{\linewidth}{!}{
\begin{tabular}{c|c|ccc|ccc|ccc} 
\toprule
\multirow{2}{*}{\textbf{Methods}} & \multirow{2}{*}{\textbf{Dataset}} & \multicolumn{3}{c|}{\textbf{Hausdorff}}                                                                   & \multicolumn{3}{c|}{\textbf{Fréchet}}                                                                     & \multicolumn{3}{c}{\textbf{DTW}}                                                                           \\ 
\cmidrule(l){3-11}
                                  &                                   & \textbf{HR@10}                    & \textbf{HR@50}                    & \textbf{R10@50}                   & \textbf{HR@10}                    & \textbf{HR@50}                    & \textbf{R10@50}                   & \textbf{HR@10}                    & \textbf{HR@50}                    & \textbf{R10@50}                    \\ 
\midrule
PCA        \cite{wold1987principal}                       & \multirow{8}{*}{Porto}            & 0.4850                            & 0.5439                            & 0.8454                            & 0.4203                            & 0.4909                            & 0.8038                            & 0.4751                            & 0.5746                            & 0.8534                             \\
SVD           \cite{wall2003singular}                    &                                   & 0.4839                            & 0.5436                            & 0.8445                            & 0.4294                            & 0.4977                            & 0.8106                            & 0.4689                            & 0.5591                            & 0.8362                             \\
MDS            \cite{kruskal1978multidimensional}                   &                                   & 0.4839                            & 0.6065                            & 0.8770                            & 0.4661                            & 0.5874                            & 0.8607                            & 0.4762                            & 0.5865                            & 0.8673                             \\
Siamese     \cite{pei2016modeling}                      &                                   & 0.3834                            & 0.4999                            & 0.7760                            & 0.4740                            & 0.5802                            & 0.7970                            & 0.3832                            & 0.4804                            & 0.7602                             \\
NEUTRAJ        \cite{yao2019computing}                   &                                   & 0.4372                            & 0.5714                            & 0.8089                            & 0.5225                            & 0.6351                            & 0.8292                            & 0.4370                            & 0.5613                            & 0.8396                             \\
T3S         \cite{yang2021t3s}                      &                                   & 0.4672                            & 0.5977                            & 0.8344                            & 0.5518                            & \textcolor{blue}{\textbf{0.6560}} & 0.8550                            & 0.4345                            & 0.5809                            & 0.8350                             \\ 
\cmidrule{1-1}\cmidrule{3-11}
\textbf{Efflex-B (Ours)}          &                                   & \textcolor{blue}{\textbf{0.5510}} & \textcolor{blue}{\textbf{0.6492}} & \textcolor{blue}{\textbf{0.9817}} & \textcolor{blue}{\textbf{0.5564}} & 0.6273                            & \textcolor{blue}{\textbf{0.9750}} & \textcolor{blue}{\textbf{0.5760}} & \textcolor{blue}{\textbf{0.5892}} & \textcolor{blue}{\textbf{0.9080}}  \\
\textbf{Efflex-L (Ours)}          &                                   & \textbf{\textcolor{red}{0.5651}}  & \textbf{\textcolor{red}{0.7126}}  & \textbf{\textcolor{red}{0.9984}}  & \textbf{\textcolor{red}{0.5705}}  & \textbf{\textcolor{red}{0.7139}}  & \textbf{\textcolor{red}{0.9984}}  & \textbf{\textcolor{red}{0.6412}}  & \textbf{\textcolor{red}{0.7195}}  & \textbf{\textcolor{red}{0.9965}}   \\ 
\midrule
PCA    \cite{wold1987principal}                           & \multirow{8}{*}{Geolife}          & 0.4110                            & 0.5562                            & 0.8243                            & 0.4336                            & 0.5880                            & 0.8446                            & 0.4331                            & 0.5481                            & 0.8190                             \\
SVD        \cite{wall2003singular}                       &                                   & 0.4081                            & 0.5563                            & 0.8248                            & 0.4438                            & 0.6041                            & 0.8448                            & 0.4481                            & 0.5285                            & 0.8133                             \\
MDS         \cite{kruskal1978multidimensional}                      &                                   & 0.3602                            & 0.5472                            & 0.8535                            & 0.4793                            & 0.6187                            & 0.8716                            & 0.4656                            & 0.5347                            & 0.8354                             \\
Siamese            \cite{pei2016modeling}               &                                   & 0.3120                            & 0.4236                            & 0.6640                            & 0.4631                            & 0.6032                            & 0.8121                            & 0.2680                            & 0.4582                            & 0.6172                             \\
NEUTRAJ      \cite{yao2019computing}                     &                                   & 0.3691                            & 0.4870                            & 0.7416                            & 0.4947                            & \textbf{\textcolor{blue}{0.6786}} & 0.8403                            & 0.3067                            & 0.4832                            & 0.6513                             \\
T3S         \cite{yang2021t3s}                      &                                   & 0.3807                            & 0.5463                            & 0.7690                            & 0.5231                            & 0.6732                            & 0.8667                            & 0.3208                            & 0.4316                            & 0.6601                             \\ 
\cmidrule{1-1}\cmidrule{3-11}
\textbf{Efflex-B (Ours)}          &                                   & \textcolor{blue}{\textbf{0.5621}} & \textcolor{blue}{\textbf{0.6464}} & \textcolor{blue}{\textbf{0.9694}} & \textcolor{blue}{\textbf{0.5828}} & 0.6439                            & \textbf{\textcolor{blue}{0.9601}} & \textbf{\textcolor{blue}{0.6034}} & \textbf{\textcolor{blue}{0.6271}} & \textbf{\textcolor{blue}{0.9165}}  \\
\textbf{Efflex-L (Ours)}          &                                   & \textbf{\textcolor{red}{0.6030}}  & \textbf{\textcolor{red}{0.7303}}  & \textbf{\textcolor{red}{0.9929}}  & \textbf{\textcolor{red}{0.6163}}  & \textbf{\textcolor{red}{0.7425}}  & \textbf{\textcolor{red}{0.9947}}  & \textbf{\textcolor{red}{0.6975}}  & \textbf{\textcolor{red}{0.7706}}  & \textbf{\textcolor{red}{0.9970}}   \\
\bottomrule
\end{tabular}
}
\caption{\textbf{Quantitative performance compared with state-of-the-art models on graph representation learning.} We evaluate under three distance functions (Fréchet, Hausdorff, and DTW) with multiple evaluation metrics (hitting ratio: HR@10, HR@50, recall: R10@50) employed for each distance under two datasets: Porto and Geolife. 
\textcolor{red}{\textbf{Red}} / \textcolor{blue}{\textbf{Blue}} numbers: Highest/Second highest among all methods.
}
\label{bigtable}
\end{table*}

\subsection{Efflex with Flexibility}
Benefiting from the generalized and flexible pipeline framework, Efflex offers two versions, Efflex-B and Efflex-L, with different models employed in the Graph Representation Learning Module. 
Specifically, Efflex-B uses the proposed lightweight GCN for representation learning, which achieves accurate and competitive accuracy while improving the training speed significantly ($\times$ 36 faster). 
Meanwhile, we replace the lightweight GCN with the deepwalk-based node2vec \cite{grover2016node2vec} model with massive parameters to learn graph embeddings, and regard the new version as Efflex-L. As shown in Section \ref{sec:exp}, Efflex-L reaches state-of-the-art performance under various evaluation metrics. 

The two versions offered by the Efflex pipeline (Efflex-B/L) focus on diverse application needs. The base version specializes in applications requiring real-time modeling including wearable devices and embedded systems, while the large version can be used for environmental monitoring and smart city infrastructure management.
Section \ref{subsec:effana} analyzes the performance of these two versions.

\section{Experiments}
\label{sec:exp}

\begin{figure*}[htbp]
  \centering
    \includegraphics[width=0.9\textwidth]{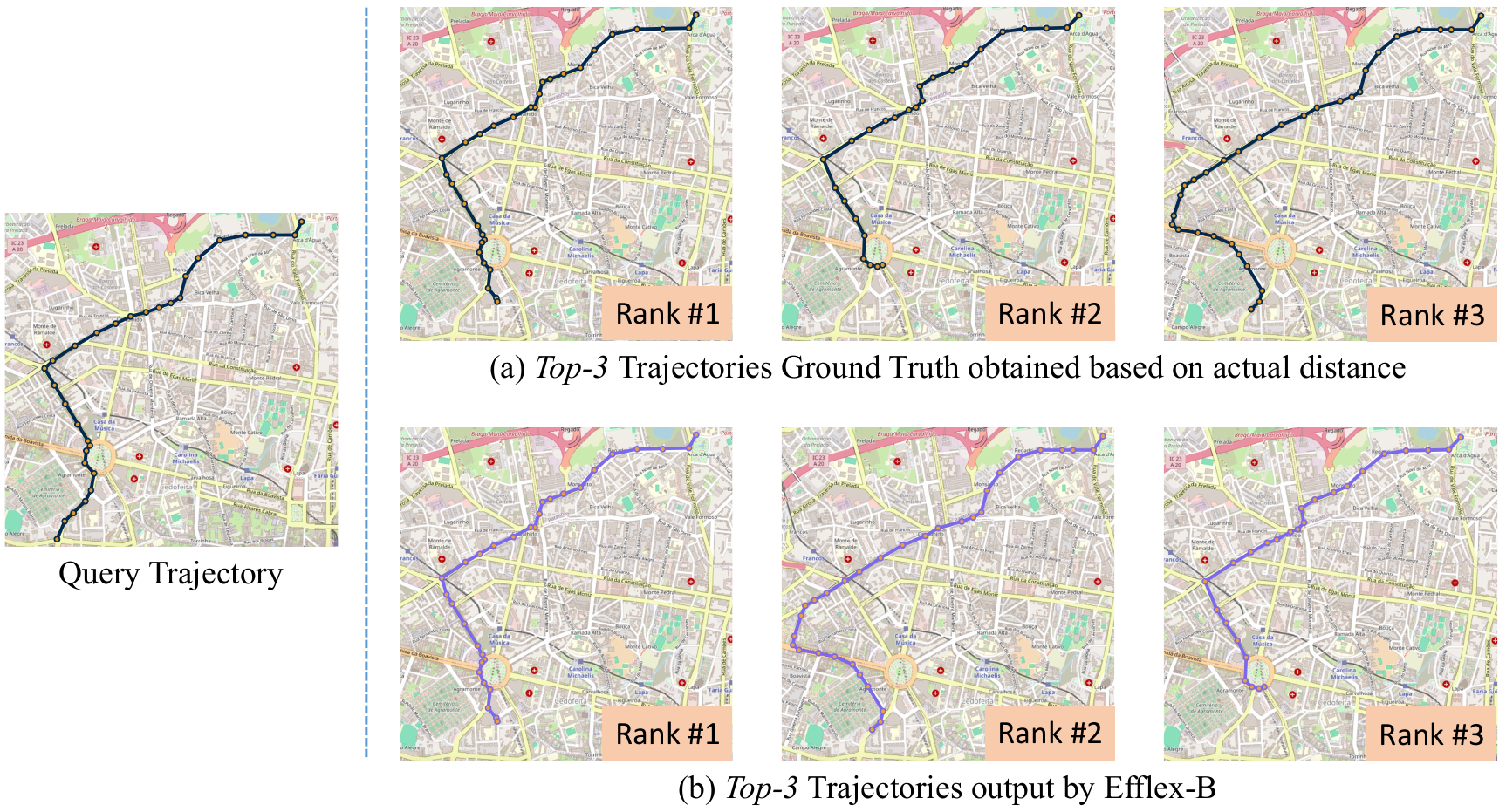}
    \caption{
    \textbf{Qualitative visualization of the trajectory similarity search task.}
    \textbf{Left:} Query trajectory.
    \textbf{Right (a):} \textit{Top-3} similar ground truth trajectories.
    \textbf{Right (b):} \textit{Top-3} similar retrieval results of our model (Efflex-B).
    Our retrieval results are consistent with ground truth. 
    }
  \label{fig:query}
\end{figure*}

\begin{table*}[htb]
\centering
\begin{tabular}{@{}c|c|cc|cc|cc@{}}
\toprule
\multirow{2}{*}{\textbf{Methods}} & \multirow{2}{*}{\textbf{Dataset}} & \multicolumn{2}{c|}{\textbf{Hausdorff}}                                                                                                                            & \multicolumn{2}{c|}{\textbf{Fréchet}}                                                                                                                              & \multicolumn{2}{c}{\textbf{DTW}}                                                                                                                                   \\ \cmidrule(l){3-8} 
                                  &                                   & \multicolumn{1}{c|}{\textbf{Recall}}                                                       & \textbf{Time / s}                                                     & \multicolumn{1}{c|}{\textbf{Recall}}                                                       & \textbf{Time / s}                                                     & \multicolumn{1}{c|}{\textbf{Recall}}                                                       & \textbf{Time / s}                                                     \\ \midrule
\textbf{Efflex-L}                 & \multirow{3}{*}{{Porto}}   & \multicolumn{1}{c|}{0.9984}                                                                & 1913.48                                                               & \multicolumn{1}{c|}{0.9984}                                                                & 1819.01                                                               & \multicolumn{1}{c|}{0.9965}                                                                & 1382.78                                                               \\
\textbf{Efflex-B}                 &                                   & \multicolumn{1}{c|}{0.9817}                                                                & 51.61                                                                 & \multicolumn{1}{c|}{0.9750}                                                                & 51.60                                                                 & \multicolumn{1}{c|}{0.9080}                                                                & 52.76                                                                 \\
\textbf{Diff (B vs. L)}           &                                   & \multicolumn{1}{c|}{\textcolor{blue}{$\downarrow$\textbf{1.67\%}}} & \textbf{\textcolor{red}{ $\uparrow\times$36}} & \multicolumn{1}{c|}{\textbf{\textcolor{blue}{$\downarrow$2.34\%}}} & \textbf{\textcolor{red}{$\uparrow\times$34}} & \multicolumn{1}{c|}{\textbf{\textcolor{blue}{$\downarrow$8.85\%}}} & \textbf{\textcolor{red}{$\uparrow\times$25}} \\ \midrule
\textbf{Efflex-L}                 & \multirow{3}{*}{{Geolife}} & \multicolumn{1}{c|}{0.9929}                                                                & 1309.80                                                               & \multicolumn{1}{c|}{0.9947}                                                                & 1307.65                                                               & \multicolumn{1}{c|}{0.9970}                                                                & 1263.46                                                               \\
\textbf{Efflex-B}                 &                                   & \multicolumn{1}{c|}{0.9694}                                                                & 78.89                                                                 & \multicolumn{1}{c|}{0.9601}                                                                & 77.94                                                                 & \multicolumn{1}{c|}{0.9165}                                                                & 69.28                                                                 \\
\textbf{Diff (B vs. L)}           &                                   & \multicolumn{1}{c|}{\textbf{\textcolor{blue}{$\downarrow$2.35\%}}} & \textbf{\textcolor{red}{$\uparrow\times$16}} & \multicolumn{1}{c|}{\textbf{\textcolor{blue}{$\downarrow$3.46\%}}} & \textbf{\textcolor{red}{$\uparrow\times$16}} & \multicolumn{1}{c|}{\textbf{\textcolor{blue}{$\downarrow$8.05\%}}} & \textbf{\textcolor{red}{$\uparrow\times$17}} \\ \bottomrule
\end{tabular}
\caption{
\textbf{Efficiency analysis.} 
We compare the recall and time cost (CPU) for pipeline training between Efflex-B/L under three distance functions on Porto and Geolife datasets. Efflex-B (with GCN) significantly improves the speed (up to $\times$36 faster) while maintaining a competitive accuracy against Efflex-L (with node2vec \cite{grover2016node2vec}). 
}
\label{table:eff}
\end{table*}

\subsection{Dataset}
We conduct extensive experiments on commonly used trajectory datasets collected from real-world data points -- Porto \cite{moreira2016time} and Geolife \cite{zheng2010geolife}. 

Porto \cite{moreira2016time} contains 1,704,759 taxi trajectories gathered between 2013 and 2014 within Porto, Portugal. The recorded data encompass longitude coordinates ranging from $-8.74$ to $-8.16$ and latitude coordinates spanning from $40.95$ to $41.31$. 
Similarly, the Geolife dataset \cite{zheng2010geolife} offers a rich collection of GPS trajectories, capturing the movements of 182 users over five years (from April 2007 to August 2012). This dataset includes over 24,876 trajectories, which amounts to more than 1.2 million kilometers and a cumulative duration exceeding 48,000 hours. The geographical scope of this data spans several cities in China, with a longitude ranging from $115.9$ to $117.1$ and a latitude ranging from $39.6$ to $40.7$. 

Considering the complexity and variability observed in real-world traffic trajectories, the Porto and Geolife datasets are ideal for evaluating model performance.

\subsection{Implementation Details}

The datasets are preprocessed by excluding trajectories with fewer than 50 data points. Then we partition the dataset into 50m$\times$50m grids, following the standard operation \cite{yao2019computing}. 
We set the initial learning rate as 0.001, and use StepLR as the learning rate scheduler, which decreases by a factor of 0.1 every 5 epochs. The total training epochs are 50. The model's parameters are updated through the AdamW optimizer. 
Our experiments are conducted on AMD EPYC 7313 16-Core CPU and NVIDIA RTX A6000 GPU. 
To ensure a comprehensive assessment of runtime efficiency, we benchmark the performance of all compared algorithms using a CPU in single-threaded mode.

\subsection{Evaluation Metrics}
To conduct objective evaluation, we evaluate the model's performance on the top-$N$ similarity search task problem, following the state-of-the-art models \cite{yao2019computing, yang2021t3s}.
Specifically, given a query trajectory, the model outputs its top-$N$ similar trajectories based on the trajectory embeddings it learns under certain distance measurement function. 
The higher similarity search accuracy indicates the more accurate learned embeddings of the original trajectory dataset by the model. 

Following the standard evaluation procedure \cite{yao2019computing, yang2021t3s}, two evaluation metrics are involved: hitting ratio (HR@10, HR@50) and recall (R10@50). 
We compare the results with both non-learning-based methods (PCA \cite{wold1987principal}, SVD \cite{wall2003singular}, MDS \cite{kruskal1978multidimensional}) and state-of-the-art machine learning models (Siamese \cite{pei2016modeling}, NEUTRAJ \cite{yao2019computing}, T3S \cite{yang2021t3s}). 

\subsection{Quantitative Results}
\subsubsection{Representation Learning Performance}
Table \ref{bigtable} shows the graph representation learning performance, where both our two solutions (base and large version) outperform all existing models and achieve the state-of-the-art.
As mentioned previously, Efflex-B utilizes the lightweight GCN for graph representation learning, while Efflex-L employs node2vec \cite{grover2016node2vec} with massive trainable parameters for extracting graph embeddings.  

Specifically, under Hausdorff distance, Efflex-B reaches the hitting ratio and recall of $55.10\%$, $64.92\%$, and $98.17\%$, while Efflex-L achieves a more accurate result: $56.51\%$, $71.26\%$, and $99.84\%$. 
In terms of the comparison with matrix factorization-based methods, Efflex-B/L demonstrates significant improvement over PCA ($+10.53/16.87\%$), SVD ($+10.56/16.90\%$), and MDS ($+4.27/10.61\%$) of HR@50. Similar results can be observed for other metrics (HR@10, R10@50).
Similar results can be observed from Geolife.

Moreover, when compared with learning-based methods, Efflex-B/L consistently showcases its effectiveness, with a remarkable lead (HR@50 as an example) of $+14.93/21.27\%$, $+7.78/14.12\%$, and $+5.15/11.49\%$ against Siamese \cite{pei2016modeling}, NEUTRAJ \cite{yao2019computing}, and T3S \cite{yang2021t3s}, respectively. Similar observations can be found in Geolife. Benefiting from the design of multi-scale KNN graph construction and fusion mentioned in Section \ref{sec:adjmatrix}, our model can selectively preserve the important global and local features when transforming the original trajectory dataset into the relatively low-dimension graph. In addition, the utilization of GCN allows our model to eventually converge for learning graph structures and capturing graph-level representations, leading to significant improvements compared with RNN \cite{pei2016modeling} / LSTM \cite{yao2019computing, yang2021t3s} based methods. 

Similar observations can be concluded from Table \ref{bigtable} under either Fréchet or DTW measurement, proving the consistent effectiveness and robustness of our model across different evaluation metrics. 

\begin{figure*}[htbp]
  \centering
    \includegraphics[width=\textwidth]{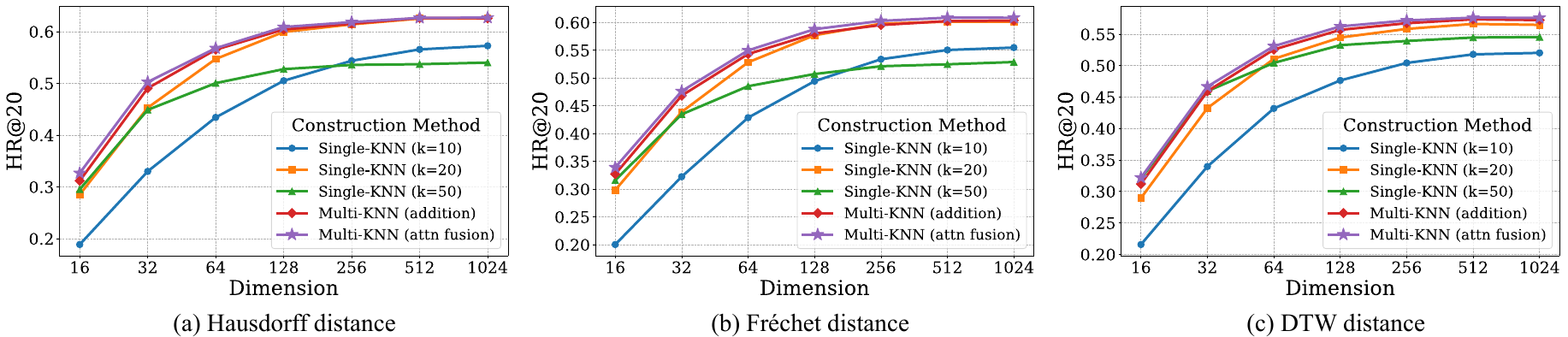}
    \caption{
    \textbf{Ablation study on different structures and embedding dimensions under three distances on Porto.}
    We compare the results of single-scale KNN ($k$=10,20,50), multi-scale KNN (with different fusion strategies), and different output embedding dimensions. 
    }
  \label{fig:ablation_curve}
\end{figure*}

\subsubsection{Efficiency Analysis}
\label{subsec:effana}
We compare the efficiency of the two versions of our model, Efflex-B/L, on Porto and Geolife datasets, as shown in Table \ref{table:eff}, where both models are evaluated under the same CPU environment. We utilize recall (R10@50) as the embedding learning accuracy metric and compute the time cost under three distance functions. 
From Table \ref{table:eff}, it is evidently observed that Efflex-B can reach a competitive accuracy while significantly reducing the time cost. Specifically, under Hausdorff distance, although Efflex-B slightly reduces the accuracy by $1.67\%$, it reaches the speed that is 36 times faster. 
For Fréchet distance, Efflex-B consistently showcases its effectiveness and efficiency, with an accuracy deducted by $2.34\%$, it also improves an extraction speed by 34 times faster.  Similar results can be observed from DTW distance and in the Geolife dataset.

Benefiting from our lightweight but effective design of GCN, Efflex-B can effectively capture graph structure patterns for accurate representation learning while noticeably improving the training speed, compared with Efflex-L with node2vec \cite{grover2016node2vec} as the representation learning backbone. 
Since Efflex-B is a solely innovative design (without involving existing architecture \cite{grover2016node2vec} as Efflex-L), it strongly showcases our contributions to effectively learning graph representations on spatio-temporal data.  

Moreover, the substantial improvement in modeling accuracy demonstrates the potential of Efflex-B for future applications on lightweight platforms such as mobile phones and wearable devices, and autonomous driving. 
As for application scenarios that do not require real-time modeling including basic environmental monitoring, our Efflex-L version can be widely used, considering its state-of-the-art learning accuracy.

\subsection{Qualitative Results}
The qualitative demonstration is illustrated in Figure \ref{fig:query}, 
which shows the trajectory query results (Porto as an example). Specifically, given a query trajectory, the model will find the \textit{top-3} similar trajectories based on the learned embeddings. The corresponding ground truth is retrieved by computing the actual distance in the dataset. 
From Figure \ref{fig:query},
we can observe that the retrieval results of Efflex closely match the ground truth (i.e., Rank \#1 in Ground Truth vs. Rank \#1 in our results), proving the effectiveness of our model on the trajectory similarity search task. Meanwhile, the convincing results justify the theory of dimensionalizing the original spatio-temporal trajectories for graph construction, which provides an applicable solution to represent the massive spatio-temporal data efficiently. 
Additionally, the potential real-life application scenarios (i.e., finding alternative trajectories when the existing path cannot be traveled when navigating) show Efflex's application and generalization.

\subsection{Ablation Studies}
\label{subsection:ablation}
We conduct comprehensive ablation studies to verify the effectiveness of the model's key components and parameters, including attention for feature fusion, multi-scale graph construction, output embedding dimension, and loss terms.

\begin{table}[htbp]
\centering
\resizebox{1.0\columnwidth}{!}{%
\begin{tabular}{@{}c|ccc@{}}
\toprule
\textbf{Methods}         & \textbf{Hausdorff} & \textbf{Fréchet} & \textbf{DTW} \\ \midrule
w/o Multi-KNN   & 0.9723                & 0.9447              & 0.8724          \\
w/o Attn Fusion & 0.9789                & 0.9733              & 0.9023          \\ \midrule
L1 Loss         & 0.9646                & 0.9632              & 0.8107          \\
MSE Loss        & 0.9698                & 0.9658              & 0.8094          \\ \midrule
\textbf{Efflex (Ours)}         & \textbf{0.9817}       & \textbf{0.9750}     & \textbf{0.9080} \\ \bottomrule
\end{tabular}
}
\caption{
\textbf{Ablation studies on model structure and different losses under three distances on Porto.} 
Our final design showcases the most accurate embedding learning of the graph. 
}
\label{table:ablation}
\end{table}

\noindent
\textbf{Attention for Feature Fusion.} The ablation study on the proposed multi-scale KNN graph construction and fusion is shown in Figure \ref{fig:ablation_curve}, where we report the hitting ratio score against embedding dimensions under three distance functions. 
Specifically, we compare the results of the following: multi-scale graph construction and fusion with attention (final model), multi-scale construction and fusion through simple addition operation, and single-scale graph construction ($k$ = 10, 20, and 50).

It is evident that our design (multi-scale KNN with attention) outperforms other structures for different embedding dimensions under all distance functions, with the purple line consistently superior to others.  
Since the proposed lightweight attention allows the model to selectively and dynamically capture the important features and graph structure information with different scales (global and local levels), it contributes to more accurate embedding learning than other simple fusion strategies (addition operation), with the purple line is higher than the red one. 
The quantitative analysis on attention fusion is demonstrated in Table \ref{table:ablation}, where ``w/o Attn Fusion'' refers to utilizing the simple addition operation to fuse features. 

\noindent
\textbf{Multi-scale vs. Single-scale Graph Construction.}
In Figure \ref{fig:ablation_curve}, both purple and red lines remain at the highest levels under different conditions, indicating the effectiveness of multi-scale graph construction over single-scale one. The employment of the KNN algorithm with different scales allows the model to capture board relationships and localized patterns at the same time, leading to a more comprehensive understanding of the graph structure compared with single-scale KNN ($k$ = 10, 20, and 50).  
The quantitative result is shown in Table \ref{table:ablation}, where ``w/o Multi-KNN'' indicates the single-scale KNN algorithm ($k$ = 20 as an example). 

\noindent
\textbf{Output Embedding Dimension.} 
Since the original trajectory is transferred into low-level embeddings, different embedding dimensions determine the representation quality. We conduct the ablation analysis of different embedding dimensions ranging from 16 to 1024, as shown in Figure \ref{fig:ablation_curve}.
As the embedding dimension increases, the model's performance shows a significant increase followed by a gradually stable trend. Increasing the dimension of the embeddings allows the representation space to be closer to the original high-dimensional space (spatio-temporal trajectory data), assisting the model in better representing the original data. 

\noindent
\textbf{Loss Terms.} 
The result of different losses for pipeline training is shown in Table \ref{table:ablation}, where cosine similarity distance outperforms L1 Loss and MSE Loss. 
Considering the ability of cosine similarity distance to ignore the effect of data sparsity and dimensionality \cite{yu2020semantic, yao2020cosine}, it is 
less sensitive to outliers and variations in the magnitude of the spatio-temporal graph representation learning. This indicates it to be the ideal loss term for stable pipeline training.

\section{Future Work}

In the future, we plan to focus on optimizing the framework for even greater scalability and under real-time scenario settings. This would be particularly beneficial for applications requiring immediate insights from vast amounts of spatio-temporal data, such as traffic management systems and real-time environmental monitoring.

Meanwhile, another interesting direction we are looking into involves integrating Large Language Models (LLMs) as encoders to process graph features within the Efflex pipeline, as LLM holds the potential to conveniently capture the complex semantics of spatio-temporal data, translating it into richer, context-aware embeddings. Such an integration could enhance the predictive accuracy and analytical depth of the Efflex framework, opening new avenues in spatio-temporal data analytics.
\section{Conclusion}
\label{sec:conclusion}
 In this paper, we introduce a novel framework that addresses the challenges of effectively learning representations from large-volume spatio-temporal trajectory data. Our comprehensive pipeline, Efflex, integrates a multi-scale KNN algorithm with feature fusion for graph construction, achieving significant advancements in dimensionality reduction while preserving essential data features. Furthermore, our custom-built lightweight GCN enhances the model's efficiency, enhancing the embedding extraction speed by up to 36 times faster without compromising accuracy.

 We demonstrate Efflex's superior performance through extensive experimens with the Porto and Geolife datasets, establishing new benchmarks in the domain. Efflex is presented in two versions, Efflex-B and Efflex-L, tailored to scenarios demanding high accuracy and environments requiring swift data processing, respectively. This dual-version approach highlights our framework's adaptability and broad applicability, underscoring its potential in time-sensitive and computationally constrained applications.

{
    \small
    \bibliographystyle{ieeenat_fullname}
    \bibliography{main}

\begin{thebibliography}{53}
\providecommand{\natexlab}[1]{#1}
\providecommand{\url}[1]{\texttt{#1}}
\expandafter\ifx\csname urlstyle\endcsname\relax
  \providecommand{\doi}[1]{doi: #1}\else
  \providecommand{\doi}{doi: \begingroup \urlstyle{rm}\Url}\fi

\bibitem[Abu-El-Haija et~al.(2021)Abu-El-Haija, Mostafa, Nassar, Crespi, Ver~Steeg, and Galstyan]{abu2021implicit}
Sami Abu-El-Haija, Hesham Mostafa, Marcel Nassar, Valentino Crespi, Greg Ver~Steeg, and Aram Galstyan.
\newblock Implicit svd for graph representation learning.
\newblock \emph{Advances in Neural Information Processing Systems}, 34:\penalty0 8419--8431, 2021.

\bibitem[Avalos et~al.(2018)Avalos, Nock, Ong, Rouar, and Sun]{avalos2018representation}
Marta Avalos, Richard Nock, Cheng~Soon Ong, Julien Rouar, and Ke Sun.
\newblock Representation learning of compositional data.
\newblock \emph{Advances in Neural Information Processing Systems}, 31, 2018.

\bibitem[Belogay et~al.(1997)Belogay, Cabrelli, Molter, and Shonkwiler]{belogay1997calculating}
E Belogay, C Cabrelli, U Molter, and R Shonkwiler.
\newblock Calculating the hausdorff distance between curves.
\newblock \emph{Information Processing Letters}, 64\penalty0 (1), 1997.

\bibitem[Bengio et~al.(2013)Bengio, Courville, and Vincent]{bengio2013representation}
Yoshua Bengio, Aaron Courville, and Pascal Vincent.
\newblock Representation learning: A review and new perspectives.
\newblock \emph{IEEE transactions on pattern analysis and machine intelligence}, 35\penalty0 (8):\penalty0 1798--1828, 2013.

\bibitem[Cheng et~al.(2023)Cheng, Diao, Cheng, and Liu]{cheng2023saic}
Ming Cheng, Xingjian Diao, Shitong Cheng, and Wenjun Liu.
\newblock Saic: Integration of speech anonymization and identity classification.
\newblock \emph{arXiv preprint arXiv:2312.15190}, 2023.

\bibitem[Cheng et~al.(2024)Cheng, Zhang, Wang, Zhou, Feng, Lyu, and Diao]{cheng2024vetrass}
Ming Cheng, Bowen Zhang, Ziyu Wang, Ziyi Zhou, Weiqi Feng, Yi Lyu, and Xingjian Diao.
\newblock Vetrass: Vehicle trajectory similarity search through graph modeling and representation learning, 2024.

\bibitem[Diao et~al.(2023{\natexlab{a}})Diao, Cheng, Barrios, and Jin]{diao2023ft2tf}
Xingjian Diao, Ming Cheng, Wayner Barrios, and SouYoung Jin.
\newblock Ft2tf: First-person statement text-to-talking face generation.
\newblock \emph{arXiv preprint arXiv:2312.05430}, 2023{\natexlab{a}}.

\bibitem[Diao et~al.(2023{\natexlab{b}})Diao, Cheng, and Cheng]{diao2023av}
Xingjian Diao, Ming Cheng, and Shitong Cheng.
\newblock Av-maskenhancer: Enhancing video representations through audio-visual masked autoencoder.
\newblock In \emph{2023 IEEE 35th International Conference on Tools with Artificial Intelligence (ICTAI)}, pages 354--360. IEEE, 2023{\natexlab{b}}.

\bibitem[Ding et~al.(2022)Ding, Xi, Wu, Liu, Wang, and Zhou]{ding2022analyzing}
Jiaxin Ding, Shichuan Xi, Kailong Wu, Pan Liu, Xinbing Wang, and Chenghu Zhou.
\newblock Analyzing sensitive information leakage in trajectory embedding models.
\newblock In \emph{Proceedings of the 30th International Conference on Advances in Geographic Information Systems}, pages 1--10, 2022.

\bibitem[Dziugaite and Roy(2015)]{dziugaite2015neural}
Gintare~Karolina Dziugaite and Daniel~M Roy.
\newblock Neural network matrix factorization.
\newblock \emph{arXiv preprint arXiv:1511.06443}, 2015.

\bibitem[Fr{\'e}chet(1906)]{frechet1906quelques}
Maurice Fr{\'e}chet.
\newblock Sur quelques points du calcul fonctionnel.
\newblock 1906.

\bibitem[Gold and Sharir(2018)]{gold2018dynamic}
Omer Gold and Micha Sharir.
\newblock Dynamic time warping and geometric edit distance: Breaking the quadratic barrier.
\newblock \emph{ACM Transactions on Algorithms (TALG)}, 14\penalty0 (4):\penalty0 1--17, 2018.

\bibitem[Grover and Leskovec(2016)]{grover2016node2vec}
Aditya Grover and Jure Leskovec.
\newblock node2vec: Scalable feature learning for networks.
\newblock In \emph{Proceedings of the 22nd ACM SIGKDD international conference on Knowledge discovery and data mining}, pages 855--864, 2016.

\bibitem[Hamilton et~al.(2017)Hamilton, Ying, and Leskovec]{hamilton2017inductive}
Will Hamilton, Zhitao Ying, and Jure Leskovec.
\newblock Inductive representation learning on large graphs.
\newblock \emph{Advances in neural information processing systems}, 30, 2017.

\bibitem[Han et~al.(2021)Han, Wang, Yao, Shang, and Zhang]{han2021graph}
Peng Han, Jin Wang, Di Yao, Shuo Shang, and Xiangliang Zhang.
\newblock A graph-based approach for trajectory similarity computation in spatial networks.
\newblock In \emph{Proceedings of the 27th ACM SIGKDD Conference on Knowledge Discovery \& Data Mining}, pages 556--564, 2021.

\bibitem[Hu et~al.(2020)Hu, Qiao, Cheng, Liu, and Wang]{hu2020dasgil}
Hanjiang Hu, Zhijian Qiao, Ming Cheng, Zhe Liu, and Hesheng Wang.
\newblock Dasgil: Domain adaptation for semantic and geometric-aware image-based localization.
\newblock \emph{IEEE Transactions on Image Processing}, 30:\penalty0 1342--1353, 2020.

\bibitem[Huang et~al.(2023)Huang, Li, Oh, and Kang]{huang2023lstm}
Jianying Huang, Jinhui Li, Jeill Oh, and Hoon Kang.
\newblock Lstm with spatiotemporal attention for iot-based wireless sensor collected hydrological time-series forecasting.
\newblock \emph{International Journal of Machine Learning and Cybernetics}, pages 1--16, 2023.

\bibitem[Huang et~al.(2022)Huang, Ma, Dong, Foutz, and Li]{huang2022empowering}
Zheng Huang, Jing Ma, Yushun Dong, Natasha~Zhang Foutz, and Jundong Li.
\newblock Empowering next poi recommendation with multi-relational modeling.
\newblock In \emph{Proceedings of the 45th International ACM SIGIR Conference on Research and Development in Information Retrieval}, pages 2034--2038, 2022.

\bibitem[Kipf and Welling(2016)]{kipf2016semi}
Thomas~N Kipf and Max Welling.
\newblock Semi-supervised classification with graph convolutional networks.
\newblock \emph{arXiv preprint arXiv:1609.02907}, 2016.

\bibitem[Koshkak et~al.(2024)Koshkak, Wang, Kanduri, Liljeberg, Rahmani, and Dutt]{AlikhaniKoshkak2024SEAL}
Hamidreza~Alikhani Koshkak, Ziyu Wang, Anil Kanduri, Pasi Liljeberg, Amir~M. Rahmani, and Nikil Dutt.
\newblock {SEAL: Sensing Efficient Active Learning on Wearables through Context-awareness}.
\newblock In \emph{Proceedings of the IEEE/ACM Design, Automation and Test in Europe Conference}, Spain, 2024.
\newblock DATE'24.

\bibitem[Kruskal and Wish(1978)]{kruskal1978multidimensional}
Joseph~B Kruskal and Myron Wish.
\newblock \emph{Multidimensional scaling}.
\newblock Number~11. Sage, 1978.

\bibitem[Kryszkiewicz(2014)]{kryszkiewicz2014cosine}
Marzena Kryszkiewicz.
\newblock The cosine similarity in terms of the euclidean distance.
\newblock In \emph{Encyclopedia of Business Analytics and Optimization}, pages 2498--2508. IGI Global, 2014.

\bibitem[Li et~al.(2018)Li, Zhao, Cong, Jensen, and Wei]{li2018deep}
Xiucheng Li, Kaiqi Zhao, Gao Cong, Christian~S Jensen, and Wei Wei.
\newblock Deep representation learning for trajectory similarity computation.
\newblock In \emph{2018 IEEE 34th international conference on data engineering (ICDE)}, pages 617--628. IEEE, 2018.

\bibitem[Li et~al.(2015)Li, Tarlow, Brockschmidt, and Zemel]{li2015gated}
Yujia Li, Daniel Tarlow, Marc Brockschmidt, and Richard Zemel.
\newblock Gated graph sequence neural networks.
\newblock \emph{arXiv preprint arXiv:1511.05493}, 2015.

\bibitem[Loshchilov and Hutter(2017)]{loshchilov2017decoupled}
Ilya Loshchilov and Frank Hutter.
\newblock Decoupled weight decay regularization.
\newblock \emph{arXiv preprint arXiv:1711.05101}, 2017.

\bibitem[Ma et~al.(2022)Ma, Dong, Huang, Mietchen, and Li]{ma2022assessing}
Jing Ma, Yushun Dong, Zheng Huang, Daniel Mietchen, and Jundong Li.
\newblock Assessing the causal impact of covid-19 related policies on outbreak dynamics: A case study in the us.
\newblock In \emph{Proceedings of the ACM Web Conference 2022}, pages 2678--2686, 2022.

\bibitem[Ma{\'c}kiewicz and Ratajczak(1993)]{mackiewicz1993principal}
Andrzej Ma{\'c}kiewicz and Waldemar Ratajczak.
\newblock Principal components analysis (pca).
\newblock \emph{Computers \& Geosciences}, 19\penalty0 (3):\penalty0 303--342, 1993.

\bibitem[Meiler et~al.(2001)Meiler, M{\"u}ller, Zeidler, and Schm{\"a}schke]{meiler2001generation}
Jens Meiler, Michael M{\"u}ller, Anita Zeidler, and Felix Schm{\"a}schke.
\newblock Generation and evaluation of dimension-reduced amino acid parameter representations by artificial neural networks.
\newblock \emph{Molecular modeling annual}, 7\penalty0 (9):\penalty0 360--369, 2001.

\bibitem[Mnih and Salakhutdinov(2007)]{mnih2007probabilistic}
Andriy Mnih and Russ~R Salakhutdinov.
\newblock Probabilistic matrix factorization.
\newblock \emph{Advances in neural information processing systems}, 20, 2007.

\bibitem[Moreira-Matias et~al.(2016)Moreira-Matias, Gama, Ferreira, Mendes-Moreira, and Damas]{moreira2016time}
Lu{\'\i}s Moreira-Matias, Jo{\~a}o Gama, Michel Ferreira, Jo{\~a}o Mendes-Moreira, and Luis Damas.
\newblock Time-evolving od matrix estimation using high-speed gps data streams.
\newblock \emph{Expert systems with Applications}, 44:\penalty0 275--288, 2016.

\bibitem[Pei et~al.(2016)Pei, Tax, and van~der Maaten]{pei2016modeling}
Wenjie Pei, David~MJ Tax, and Laurens van~der Maaten.
\newblock Modeling time series similarity with siamese recurrent networks.
\newblock \emph{arXiv preprint arXiv:1603.04713}, 2016.

\bibitem[Tenenbaum et~al.(2000)Tenenbaum, Silva, and Langford]{tenenbaum2000global}
Joshua~B Tenenbaum, Vin~de Silva, and John~C Langford.
\newblock A global geometric framework for nonlinear dimensionality reduction.
\newblock \emph{science}, 290\penalty0 (5500):\penalty0 2319--2323, 2000.

\bibitem[Wall et~al.(2003)Wall, Rechtsteiner, and Rocha]{wall2003singular}
Michael~E Wall, Andreas Rechtsteiner, and Luis~M Rocha.
\newblock Singular value decomposition and principal component analysis.
\newblock In \emph{A practical approach to microarray data analysis}, pages 91--109. Springer, 2003.

\bibitem[Wang et~al.(2020{\natexlab{a}})Wang, Cao, and Philip]{wang2020deep}
Senzhang Wang, Jiannong Cao, and S~Yu Philip.
\newblock Deep learning for spatio-temporal data mining: A survey.
\newblock \emph{IEEE transactions on knowledge and data engineering}, 34\penalty0 (8):\penalty0 3681--3700, 2020{\natexlab{a}}.

\bibitem[Wang et~al.(2020{\natexlab{b}})Wang, Luo, and Zhou]{wang2020guardhealth}
Ziyu Wang, Nanqing Luo, and Pan Zhou.
\newblock Guardhealth: Blockchain empowered secure data management and graph convolutional network enabled anomaly detection in smart healthcare.
\newblock \emph{Journal of Parallel and Distributed Computing}, 142:\penalty0 1--12, 2020{\natexlab{b}}.

\bibitem[Wang et~al.(2024)Wang, Yang, Azimi, and Rahmani]{wang2024differential}
Ziyu Wang, Zhongqi Yang, Iman Azimi, and Amir~M Rahmani.
\newblock Differential private federated transfer learning for mental health monitoring in everyday settings: A case study on stress detection.
\newblock \emph{arXiv preprint arXiv:2402.10862}, 2024.

\bibitem[Wold et~al.(1987)Wold, Esbensen, and Geladi]{wold1987principal}
Svante Wold, Kim Esbensen, and Paul Geladi.
\newblock Principal component analysis.
\newblock \emph{Chemometrics and intelligent laboratory systems}, 2\penalty0 (1-3):\penalty0 37--52, 1987.

\bibitem[Wu et~al.(2019)Wu, Souza, Zhang, Fifty, Yu, and Weinberger]{wu2019simplifying}
Felix Wu, Amauri Souza, Tianyi Zhang, Christopher Fifty, Tao Yu, and Kilian Weinberger.
\newblock Simplifying graph convolutional networks.
\newblock In \emph{International conference on machine learning}, pages 6861--6871. PMLR, 2019.

\bibitem[Yang et~al.(2021{\natexlab{a}})Yang, Chen, Wang, and Shang]{yang2021towards}
Chengcheng Yang, Lisi Chen, Hao Wang, and Shuo Shang.
\newblock Towards efficient selection of activity trajectories based on diversity and coverage.
\newblock In \emph{Proceedings of the AAAI Conference on Artificial Intelligence}, pages 689--696, 2021{\natexlab{a}}.

\bibitem[Yang et~al.(2021{\natexlab{b}})Yang, Wang, Zhang, Qin, Zhang, and Lin]{yang2021t3s}
Peilun Yang, Hanchen Wang, Ying Zhang, Lu Qin, Wenjie Zhang, and Xuemin Lin.
\newblock T3s: Effective representation learning for trajectory similarity computation.
\newblock In \emph{2021 IEEE 37th International Conference on Data Engineering (ICDE)}, pages 2183--2188. IEEE, 2021{\natexlab{b}}.

\bibitem[Yang et~al.(2022)Yang, Liu, Wang, and Gao]{yang2022zebra}
Xinyu Yang, Haoyuan Liu, Ziyu Wang, and Peng Gao.
\newblock Zebra: Deeply integrating system-level provenance search and tracking for efficient attack investigation.
\newblock \emph{arXiv preprint arXiv:2211.05403}, 2022.

\bibitem[Yao et~al.(2019)Yao, Cong, Zhang, and Bi]{yao2019computing}
Di Yao, Gao Cong, Chao Zhang, and Jingping Bi.
\newblock Computing trajectory similarity in linear time: A generic seed-guided neural metric learning approach.
\newblock In \emph{2019 IEEE 35th international conference on data engineering (ICDE)}, pages 1358--1369. IEEE, 2019.

\bibitem[Yao et~al.(2022)Yao, Hu, Du, Cong, Han, and Bi]{yao2022trajgat}
Di Yao, Haonan Hu, Lun Du, Gao Cong, Shi Han, and Jingping Bi.
\newblock Trajgat: A graph-based long-term dependency modeling approach for trajectory similarity computation.
\newblock In \emph{Proceedings of the 28th ACM SIGKDD conference on knowledge discovery and data mining}, pages 2275--2285, 2022.

\bibitem[Yao et~al.(2020{\natexlab{a}})Yao, Huang, Hu, and Xie]{yao2020cosine}
Huaxiong Yao, Yang Huang, Jiabei Hu, and Wenqi Xie.
\newblock Cosine similarity distance pruning algorithm based on graph attention mechanism.
\newblock In \emph{2020 IEEE International Conference on Big Data (Big Data)}, pages 3311--3318. IEEE, 2020{\natexlab{a}}.

\bibitem[Yao et~al.(2020{\natexlab{b}})Yao, Wang, and Zhou]{yao2020privacy}
Yuanfan Yao, Ziyu Wang, and Pan Zhou.
\newblock Privacy-preserving and energy efficient task offloading for collaborative mobile computing in iot: An admm approach.
\newblock \emph{Computers \& Security}, 96:\penalty0 101886, 2020{\natexlab{b}}.

\bibitem[Yin and Cui(2016)]{yin2016spatio}
Hongzhi Yin and Bin Cui.
\newblock \emph{Spatio-temporal recommendation in social media}.
\newblock Springer, 2016.

\bibitem[Yu et~al.(2020)Yu, Xia, Wang, Feng, and Li]{yu2020semantic}
Yue Yu, Tong Xia, Huandong Wang, Jie Feng, and Yong Li.
\newblock Semantic-aware spatio-temporal app usage representation via graph convolutional network.
\newblock \emph{Proceedings of the ACM on Interactive, Mobile, Wearable and Ubiquitous Technologies}, 4\penalty0 (3):\penalty0 1--24, 2020.

\bibitem[Zhang et~al.(2023{\natexlab{a}})Zhang, Zhao, Zhou, Liang, and Wang]{zhang2023doseformer}
Cao Zhang, Xiaohui Zhao, Ziyi Zhou, Xingyuan Liang, and Shuai Wang.
\newblock Doseformer: Dynamic graph transformer for postoperative pain prediction.
\newblock \emph{Electronics}, 12\penalty0 (16):\penalty0 3507, 2023{\natexlab{a}}.

\bibitem[Zhang et~al.(2021)Zhang, Feng, Li, Hou, Wang, Wang, and Guo]{zhang2021tapping}
Lu Zhang, Weiqi Feng, Chao Li, Xiaofeng Hou, Pengyu Wang, Jing Wang, and Minyi Guo.
\newblock Tapping into nfv environment for opportunistic serverless edge function deployment.
\newblock \emph{IEEE Transactions on Computers}, 71\penalty0 (10):\penalty0 2698--2704, 2021.

\bibitem[Zhang et~al.(2023{\natexlab{b}})Zhang, Li, Wang, Feng, Yu, Chen, Leng, Guo, Yang, and Yue]{zhang2023first}
Lu Zhang, Chao Li, Xinkai Wang, Weiqi Feng, Zheng Yu, Quan Chen, Jingwen Leng, Minyi Guo, Pu Yang, and Shang Yue.
\newblock First: Exploiting the multi-dimensional attributes of functions for power-aware serverless computing.
\newblock In \emph{2023 IEEE International Parallel and Distributed Processing Symposium (IPDPS)}, pages 864--874. IEEE, 2023{\natexlab{b}}.

\bibitem[Zheng et~al.(2010)Zheng, Xie, Ma, et~al.]{zheng2010geolife}
Yu Zheng, Xing Xie, Wei-Ying Ma, et~al.
\newblock Geolife: A collaborative social networking service among user, location and trajectory.
\newblock \emph{IEEE Data Eng. Bull.}, 33\penalty0 (2):\penalty0 32--39, 2010.

\bibitem[Zhou et~al.(2021)Zhou, Guo, and Zhang]{zhou2021doseguide}
Ziyi Zhou, Baoshen Guo, and Cao Zhang.
\newblock Doseguide: A graph-based dynamic time-aware prediction system for postoperative pain.
\newblock In \emph{2021 IEEE 27th International Conference on Parallel and Distributed Systems (ICPADS)}, pages 474--481. IEEE, 2021.

\bibitem[Zhu et~al.(2020)Zhu, Wang, Guo, Ling, Zhou, Tu, and He]{zhu2020sparking}
Xin Zhu, Shuai Wang, Baoshen Guo, Taiwei Ling, Ziyi Zhou, Lai Tu, and Tian He.
\newblock Sparking: A win-win data-driven contract parking sharing system.
\newblock In \emph{Adjunct Proceedings of the 2020 ACM International Joint Conference on Pervasive and Ubiquitous Computing and Proceedings of the 2020 ACM International Symposium on Wearable Computers}, pages 596--604, 2020.

\end{thebibliography}
}


\end{document}